%% file: main.tex
\documentclass[conference]{IEEEtran}
\IEEEoverridecommandlockouts

\usepackage{amsmath,amssymb,amsfonts}
\usepackage{algorithmic}
\usepackage{graphicx}
\usepackage{textcomp}
\usepackage{xcolor}
\def\BibTeX{{\rm B\kern-.05em{\sc i\kern-.025em b}\kern-.08em
    T\kern-.1667em\lower.7ex\hbox{E}\kern-.125emX}}

\usepackage{multirow}
\usepackage{booktabs} 
\usepackage[acronym]{glossaries}
\usepackage{newclude}
\usepackage{glossaries}
\usepackage{graphics}
\usepackage{graphicx}
\usepackage{wrapfig}
\usepackage{textcomp}
\usepackage[backend=biber, sorting=none]{biblatex}
\usepackage{caption, subcaption}
\usepackage[switch]{lineno} 
\usepackage{lipsum} 
\begin{document}

\newacronym{dmas}{DMAs}{District Metered Areas}
\newacronym{dma}{DMA}{District Metered Area}

\newacronym{dhn}{DHA}{District Heating Network}
\newacronym{dhs}{DHS}{District Heating Systems}
\newacronym{wdn}{WDN}{Water Distrubution Network}
\newacronym{wdns}{WDNs}{Water Distrubution Networks}
\newacronym{wdm}{WDM}{Water Demand Management}

\newacronym{cwt}{CWT}{Continous Wavelet Transform}
\newacronym{dl}{DL}{Deep Learning}
\newacronym{ml}{ML}{Machine Learning}
\newacronym{anns}{ANNs}{Artificial Neural Networks}
\newacronym{ann}{ANN}{Artificial Neural Network}
\newacronym{ai}{AI}{Artificial Intelligence}
\newacronym{rnn}{RNNs}{Recurrent Neural Networks}
\newacronym{sari}{SARIMAX}{Seasonal Auto-Regressive Integrated Moving Average with eXogenous factors}
\newacronym{cnn}{CNN}{Convolutional Neural Network}
\newacronym{DA}{da}{Day Ahead}
\newacronym{ID}{id}{Inter Day}
\newacronym{rmse}{RMSE}{Root Mean Squared Error}
\newacronym{mse}{MSE}{Mean Squared Error}
\newacronym{mae}{MAE}{Mean Absolute Error}
\newacronym{mape}{MAPE}{Mean Absolute Percentage Error}
\newacronym{smae}{SMAE}{Seasonal Mean Absolute Error}
\newacronym{mlp}{MLP}{Multi-Layer Perceptron}

\newacronym{ar}{AR}{Auto Regression}
\newacronym{ma}{MV}{Moving Averages}

\title{Water Demand Forecasting of District Metered Areas through Learned Consumer Representations
}


\author{
\IEEEauthorblockN{Adithya Ramachandran\IEEEauthorrefmark{1}, Thorkil Flensmark B. Neergaard\IEEEauthorrefmark{2}, Tomás Arias-Vergara\IEEEauthorrefmark{1},\\ Andreas Maier\IEEEauthorrefmark{1}, and Siming Bayer\IEEEauthorrefmark{1}}
\IEEEauthorblockA{\IEEEauthorrefmark{1,}Pattern Recognition Lab,
Friedrich-Alexander-University Erlangen-Nurnberg, Germany\\
Email: \IEEEauthorrefmark{1}adithya.ramachandran@fau.de}
\IEEEauthorblockA{\IEEEauthorrefmark{2}Brønderslev Forsyning A/s, Virksomhedsvej 20, Brønderslev, Denmark\\}
}

\maketitle

\input{Abstract}
\input{Introduction}

\input{Methodology}
\input{ExperimentalSetup}
\input{ResultsDiscussion}
\input{Conclusion}


\printbibliography 

\end{document}

%% file: Abstract.tex
\begin{abstract}

Advancements in smart metering technologies have significantly improved the ability to monitor and manage water utilities.
In the context of increasing uncertainty due to climate change, securing water resources and supply has emerged as an urgent global issue with extensive socioeconomic ramifications.
 Hourly consumption data from end-users have yielded substantial insights for projecting demand across regions characterized by diverse consumption patterns.
Nevertheless, the prediction of water demand remains challenging due to influencing non-deterministic factors, such as meteorological conditions.
This work introduces a novel method for short-term water demand forecasting for District Metered Areas (DMAs) which encompass commercial, agricultural, and residential consumers.
Unsupervised contrastive learning is applied to categorize end-users according to distinct consumption behaviors present within a DMA.
Subsequently, the distinct consumption behaviors are utilized as features in the ensuing demand forecasting task using wavelet-transformed convolutional networks that incorporate a cross-attention mechanism combining both historical data and the derived representations.
The proposed approach is evaluated on real-world DMAs over a six-month period, demonstrating improved forecasting performance in terms of MAPE across different DMAs, with a maximum improvement of $\mathbf{4.9\%}$. Additionally, it identifies consumers whose behavior is shaped by socioeconomic factors, enhancing prior knowledge about the deterministic patterns that influence demand.



\end{abstract}

%% file: Introduction.tex
\section{Introduction}

Water, a vital resource for life, necessitates effective water demand management to ensure ongoing and safe access. \acrfull{wdns} sustain a consistent water supply across \acrfull{dmas}, serving diverse end-consumers such as households, industries, and farms, monitored by the temporal aggregation of smart meter data \cite{distributionnetworks}. With increasing water scarcity, accurate forecasting that accounts for anthropogenic influences \cite{anthropogenic} and socio-psychographic factors \cite{COMINOLA2015198} is essential. Although aggregating demand data at the \acrshort{dma} level smoothens statistics and highlights broad consumption trends (e.g., daily and weekly seasonality), it can mask unique consumption behaviors and individual variabilities \cite{smoothing}. With varying consumer behavior complicating real-time demand modeling, it is crucial to establish a level of spatial granularity or consumer categorization to capture broad and localized patterns for forecasting.

Traditionally, time series analysis has relied heavily on statistical models, which fit known distributions using parametric functions. These models, combined with techniques like seasonal decomposition and correlational analysis, clarify intrinsic and extrinsic time series dependencies and periodicities. However, recent advancements in \acrfull{dl}, particularly contrastive learning, have facilitated the non-parametric learning of complex time series data representations, enabling the discovery of meaningful patterns from previously unknown representations \cite{meng2023unsupervisedrepresentationlearningtime}. While contrastive learning has been applied to various time series data streams including electricity, traffic, and medical for tasks like clustering, and forecasting \cite{yue2022ts2vecuniversalrepresentationtime}, \cite{faucris324516196}, its potential for modeling end-consumer water demand data remains unexplored.

Water demand forecasting has progressed from statistical and hybrid models \cite{chatterjee2021predictionhouseholdlevelheatconsumptionusing} to the extensive adoption of \acrfull{rnn} and transformer-based models \cite{du2022preformerpredictivetransformermultiscale}, \cite{KARAMAZIOTIS2020588}. The representation of input time features has also expanded beyond the time domain to include the time-frequency domain via Fourier Transform (FT), and Wavelet Transforms \cite{ramachandran2023heatdemandforecastingmultiresolutional}. Wavelet scalograms, which provide frequency details along with time localization, facilitate multi-resolution signal analysis while addressing non-stationarity \cite{waveletStationarity}, and are increasingly utilized in time series applications \cite{TAN20103606}. Despite these advancements, accurately forecasting water demand remains a challenge in scenarios where consumer behaviors are not uniform. The impact of non-deterministic factors such as culture and economy complicates demand modeling, obscuring existing demand patterns and hindering efficient water supply management.

In this study, we introduce a two-stage framework for multi-step hourly water demand forecasting that leverages unique consumption behaviors identified through contrastive learning on smart meter data within a \acrshort{dma} as additional input features for forecasting. Although existing clustering methods primarily focus on urban household patterns and intra-household activities \cite{clustertime3, clustertime4, clustertime6, clustertime38}, to our knowledge, this represents the first application of an unsupervised contrastive learning-based approach to delineate water demand patterns across varied consumer sectors, including commercial, agricultural, and residential areas. The patterns identified are subsequently utilized as features with the aggregated \acrshort{dma} demand data, allowing the model to capture both broad trends and specific consumer-induced fluctuations. This approach significantly enhances forecasting accuracy, especially in districts without a dominant urban consumer demographic.

%% file: Methodology.tex
\section{Methodology} \label{sec:methods}
\begin{figure}[bhtp]
\centerline{\includegraphics[width=2.5in]{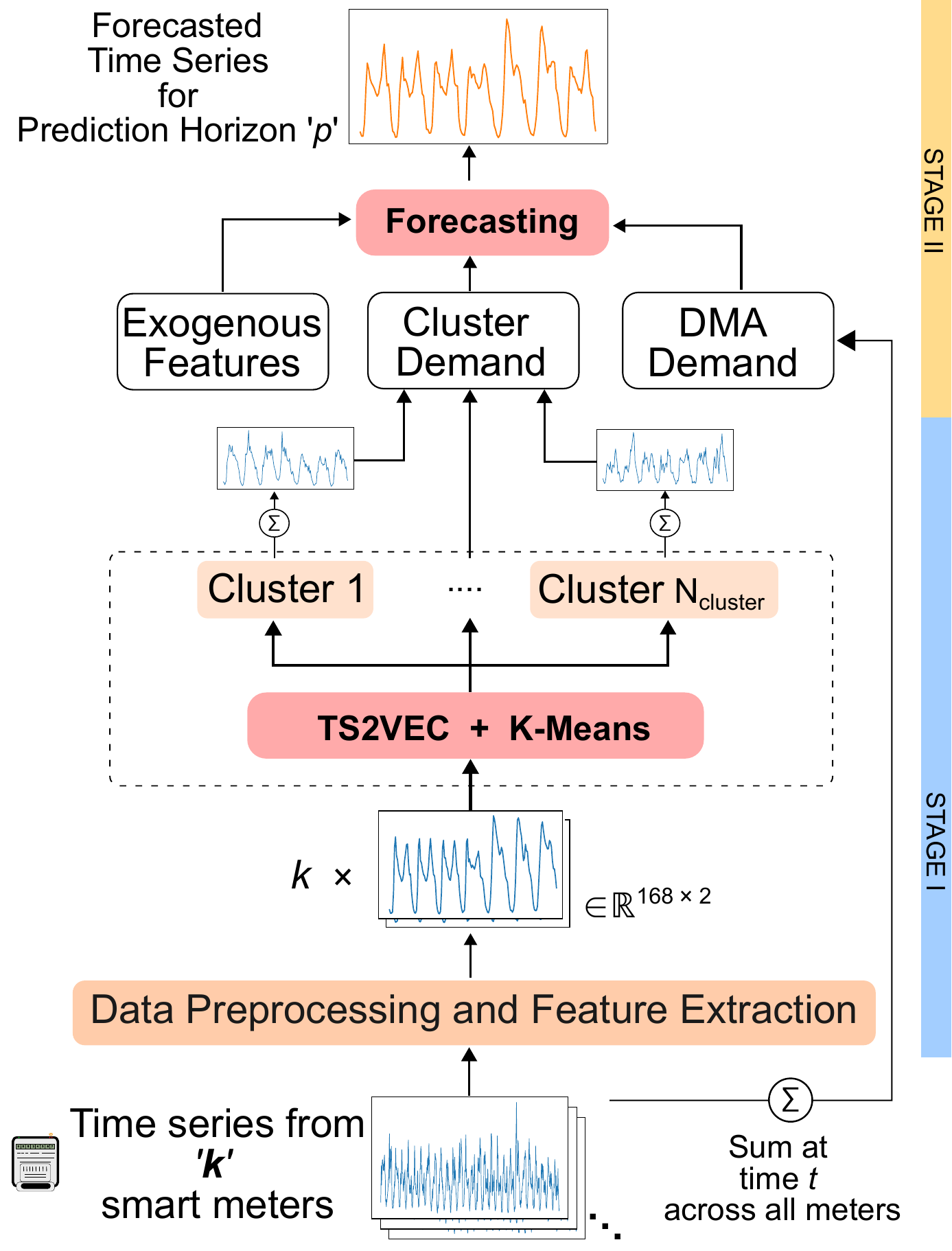}}
\caption{The proposed framework for water demand forecasting for a heterogeneous consumer \acrshort{dma}.}
\label{Fig:framework}
\end{figure}

The proposed framework is depicted in Figure \ref{Fig:framework}.
Let $ \mathbb{S} = \{  \mathbf{m}_{1}, \mathbf{m}_{2}, ..., \mathbf{m}_{k}   \} $ be the set of $k$ unique smart meters present in a \acrshort{dma}, such that $ \mathbf{m}_{i} = [  x^{i}_{1}, x^{i}_{2}, ..., x^{i}_{N}   ] \in \mathbb{R}^{N \times 1}   $, where $x^{i}_{t}$ is the recorded hourly water consumption $(m^{3}/h)$ at time $t$ by meter $ i $ over $ N $ hours. The total hourly consumption of a specific \acrshort{dma} at time $t$ is mathematically represented as $ {x}_{t} = \sum_{i=1}^{k} x_{t}^{i} $ and the downstream task of forecasting for $p$ hours is defined as the prediction $ \Tilde{\mathbf{x}} = [x_{t}, x_{t+1}, ..., x_{t+h}] $ through a deep learning model $ F $, conditioned to a set of input features $ \mathcal{X} $ consisting of historical consumption data $ [x_{t-h}, x_{t-h+1}, ..., x_{t-1}] $, where $h$ is the historical context length, and other exogenous variables of weather, time, and demographic information.

\subsection{Consumer Clustering based on Contrastive Learning}

To separate consumers with varying consumption behaviors, we adopt a contrastive learning framework TS2VEC \cite{yue2022ts2vecuniversalrepresentationtime}, that contrasts temporally within each input sequence of $\mathbf{m}_{i}$ and instance-wise across different input sequences of $\mathbf{m}_{i}$ and $\mathbf{m}_{j}$ where $i \neq j; 1 \leq i,j \leq k$ for a given batch during training.
The individual smart meter consumption data $ \mathbf{m}_{i} $ is processed to extract input samples $ \mathbf{S} \in \mathbb{R}^{168 \times 2} $ with two features, that are representative of its global and local seasonal behavior.
The first feature is a weekly load profile of $ \mathbf{m}_{i} $ obtained through averaging hourly consumption data across all weeks over the entire time period of the data.
This feature acts as an anchor for the meter's global characteristics and helps find  similarities and dissimilarities across different meters.
The second feature is also a load profile of $ \mathbf{m}_{i} $, albeit for a shorter period of 12 weeks, to capture the behavior of $ \mathbf{m}_{i} $ across different seasons of the year, and reduce the effect of stochastic temporal behavior on temporal contrasts. 
To capture the contrasting behavior of each consumer between weekdays and weekends, as well as across different consumer types, the required positive and negative pairs for contrastive learning includes temporal values representing consumption on both weekdays and weekends.
Input features from the same consumer make up the positive pair, while negative pairs are from different users. 
The hierarchical loss function employed in TS2VEC also allows for a weighting factor $\alpha$ to emphasize the contrast between samples evenly or unevenly. 
The learned representations of end-consumers are clustered through K-Means clustering \cite{kmeans++} to obtain $ N_{cluster}  $  consumer clusters with distinctive consumption characteristics and for a given cluster, the hourly demand of the cluster is the aggregation over all smart meters present in the cluster. 


\begin{figure}[bhtp]
\centerline{\includegraphics[width=2.5in]{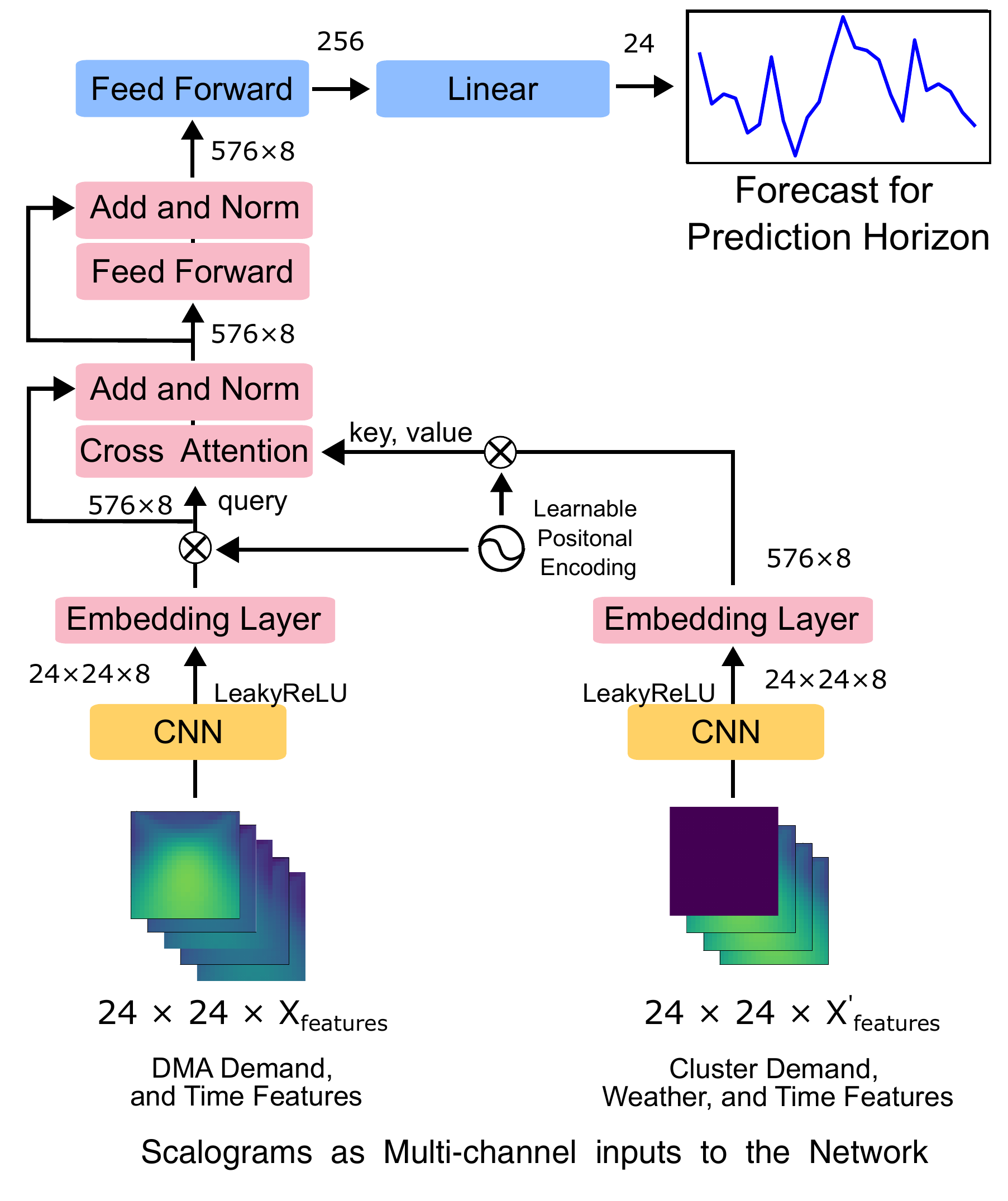}}
\caption{The architecture of the deep learning model for water demand forecasting.}
\label{Fig:forecasting}
\end{figure}

\subsection{Short-term Forecasting using Wavelet Scalograms}

To improve water demand forecasting for heterogeneous \acrshort{dma}, we employ a \acrfull{cnn} based architecture as a baseline \cite{ramachandran2023heatdemandforecastingmultiresolutional} that has been previously used for this task. The model leverages scalograms to learn feature representations from utility data.
The scalograms of $s$ scales applied at $h$ time points are obtained for $ N_{feat} $ different features and are concatenated along the third dimension, giving an image-like representation of dimensions $s \times h \times N_{feat}$, enabling 2D convolutions to learn the relationship between different features at a given time and scale $s$.
The adopted architecture is adapted optimally via a cross-attention layer \cite{faucris.324515901} to incorporate the scalograms of the obtained demand data of the clusters from the contrastive learning process.
The resulting model is $96.53\% $ smaller than the adopted baseline.
The attention block dynamically prioritizes weights to the time and frequency information present in the historical demand of the different clusters, and their contribution to the total demand of the \acrshort{dma}.
The latent representation of total DMA consumption and time features act as the query, while the latent representation of each cluster's consumption along with the exogenous factors such as maximum temperature, and humidity over the prediction horizon $p$, act as keys and values for the cross attention layer.
The two input branches along with the attention block are depicted in Figure \ref{Fig:forecasting}. The attention layer is followed by feed-forward layers with a final linear layer producing the forecasts $ \Tilde{\mathbf{y}} $ for the prediction horizon.


%% file: ExperimentalSetup.tex
\section{Experimental Setup}

\subsection{Data}
The water demand data for this retrospective analysis is from Brønderslev, Denmark and is outlined in Table \ref{Tab:data}.
Four \acrshort{dma}s including rural and semi-urban areas with heterogeneous end-users, a rural area with homogeneous end-users, and an urban \acrshort{dma} with homogeneous end-users are chosen to evaluate the effectiveness of the framework.
The urban \acrshort{dma} (predominantly residential) serves as a control to illustrate the effects of statistical smoothing during forecasting.
Geolocation of each smart meter was utilized to evaluate of clusters formed in Stage I.
The qualitative assessment across all \acrshort{dma}s featured sparse annotations by a utility company's technical expert.
Consumption data until the end of 2019 was employed for all \acrshort{dma}s, reserving the final 26 weeks (half an year) for testing, which represented between 72\% and 78\% of the total data depending on the \acrshort{dma}.
This segmentation allowed the framework to be tested across summer, winter, and a transitional season.
Moreover, 10\% of the training dataset was used as validation data. Actual recorded weather data from the respective regions are used, rather than forecasts.

\begin{table}[htbp]
\caption{Data on the selected \acrshort{dma}s for the study}
\begin{center}
\begin{tabular}{c c c c}
\hline
\textbf{ID} & \textbf{Number of meters} & \textbf{Data from} & \textbf{Description} \\
\hline
A   &  $159$ & $ 2017-09-01 $  & Rural \\  
B   & $210$ & $ 2017-09-01 $ & Semi-Urban  \\  
C   & $67$ & $ 2018-01-01 $  & Rural \\  
Control& $1286$ &   $2018-03-01 $ & Urban  \\  
\hline
\vspace{-10pt}
\end{tabular}
\label{Tab:data}
\end{center}
\end{table}

\subsection{Stage I: End-user Behaviour Clustering}

Weekly hourly load profiles (from Monday to Sunday) are utilized to learn representations of individual end-consumers.
Negative pairs consist of samples from different end-consumers, while positive pairs are derived by windowing the weekly profile of a consumer into two windows: $ w_{a} $ (Monday to Friday) and $ w_{b} $ (Friday to Sunday).
The TS2VEC model is trained with a hidden dimension of 64 and a kernel size of 3 for the 1D convolution layers in the 10 residual blocks, producing an output dimension of 16.
A custom sampler with a batch size of 64 limits each batch to only one sample from any end-consumer.
The resulting representations are clustered using K-Means, with a maximum of four clusters (residential, industrial, commercial, and agricultural).
The optimal cluster count is determined using the highest silhouette score.

\subsection{Stage II: Water Demand Forecasting}

Forecasting features for water demand are identified through correlational analysis, revealing strong correlations with historical data and moderate correlations with temperature and humidity. 
Each feature represents an observed quantity of demand, weather, or time for 24 hours.
The $X_{features}$ input features for the query branch comprise demand data with 24-hour and 168-hour lags, their seasonal components, and holiday information. 
Conversely, the $X^{'}_{features}$ input features for the key/value branch are the extracted demand from each cluster with a 24-hour lag, temperature, humidity, and sine and cosine representations of the day of the week. 
Scalograms, with $s=24$ and $h=24$, are encoded using Shannon's entropy-to-energy ratio \cite{HEIDARIBAFROUI2014437} to determine the optimal mother wavelet, identified as 'gaus4'. 

The model architecture is depicted in Figure \ref{Fig:forecasting}, detailing hidden layer output dimensions. LeakyReLU activation function is used in the convolutional and final feed-forward layers, while ReLU is employed in other feed-forward layers, with no pooling layers included.
The final block comprises hidden dimensions of 1024, 512, and 256, concluding with a linear layer for forecasting with $p=24$.
The model employs a learning rate of $0.001$, batch size of 256, ADAM optimizer, and \acrfull{mse} as the loss function, and is trained with early stopping on training and validation data set.

%% file: ResultsDiscussion.tex
\section{Results and Discussion}

\begin{figure}[bhtp]
\centerline{\includegraphics[width=3in,keepaspectratio=TRUE]{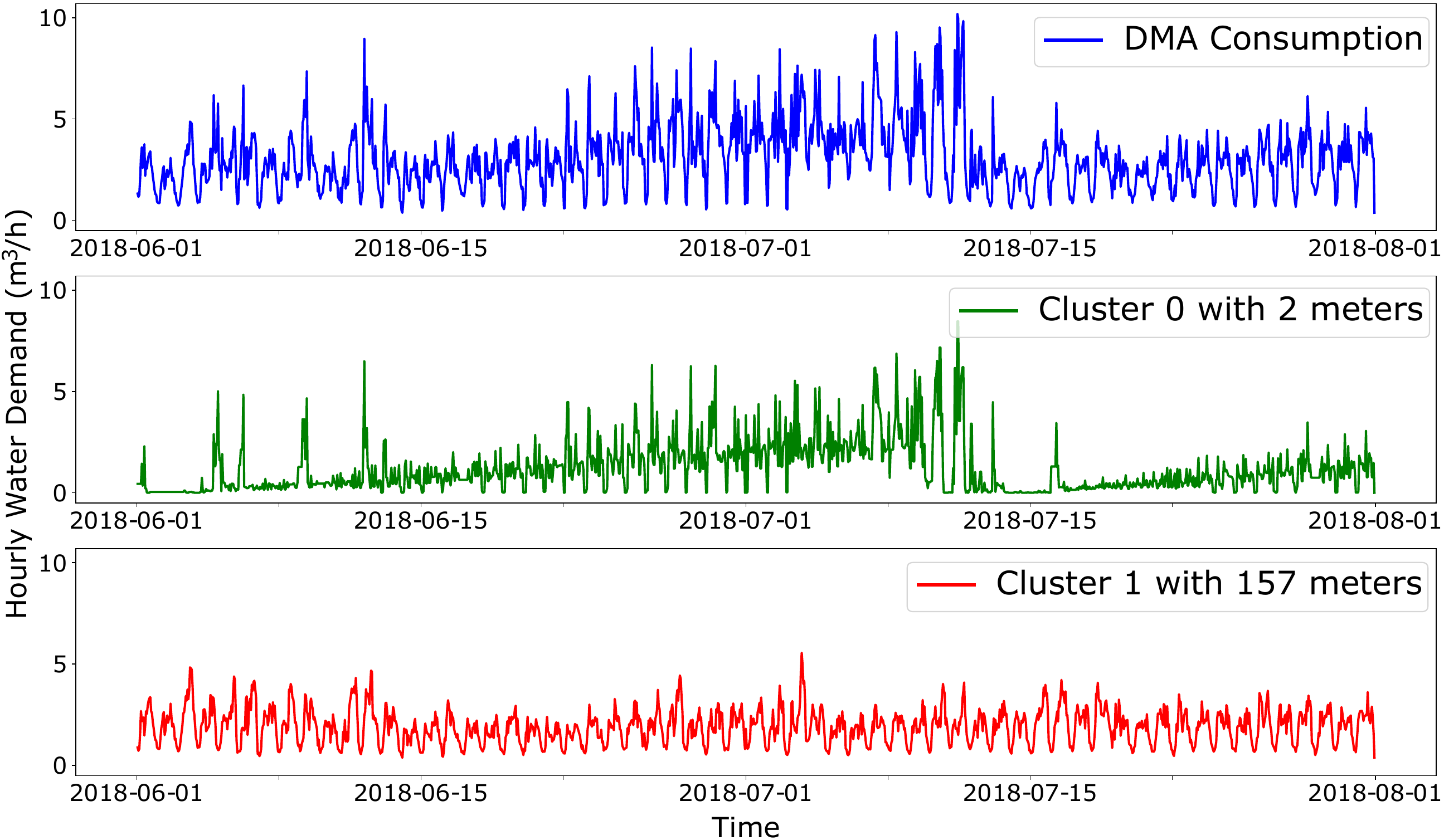}}
\caption{Observed consumption of \acrshort{dma} A along with the total demand for the obtained clusters.}
\vspace{-15pt}
\label{Fig:cluster_a}
\end{figure}

\begin{figure}[bhtp]
\centerline{\includegraphics[width=3in,keepaspectratio=TRUE]{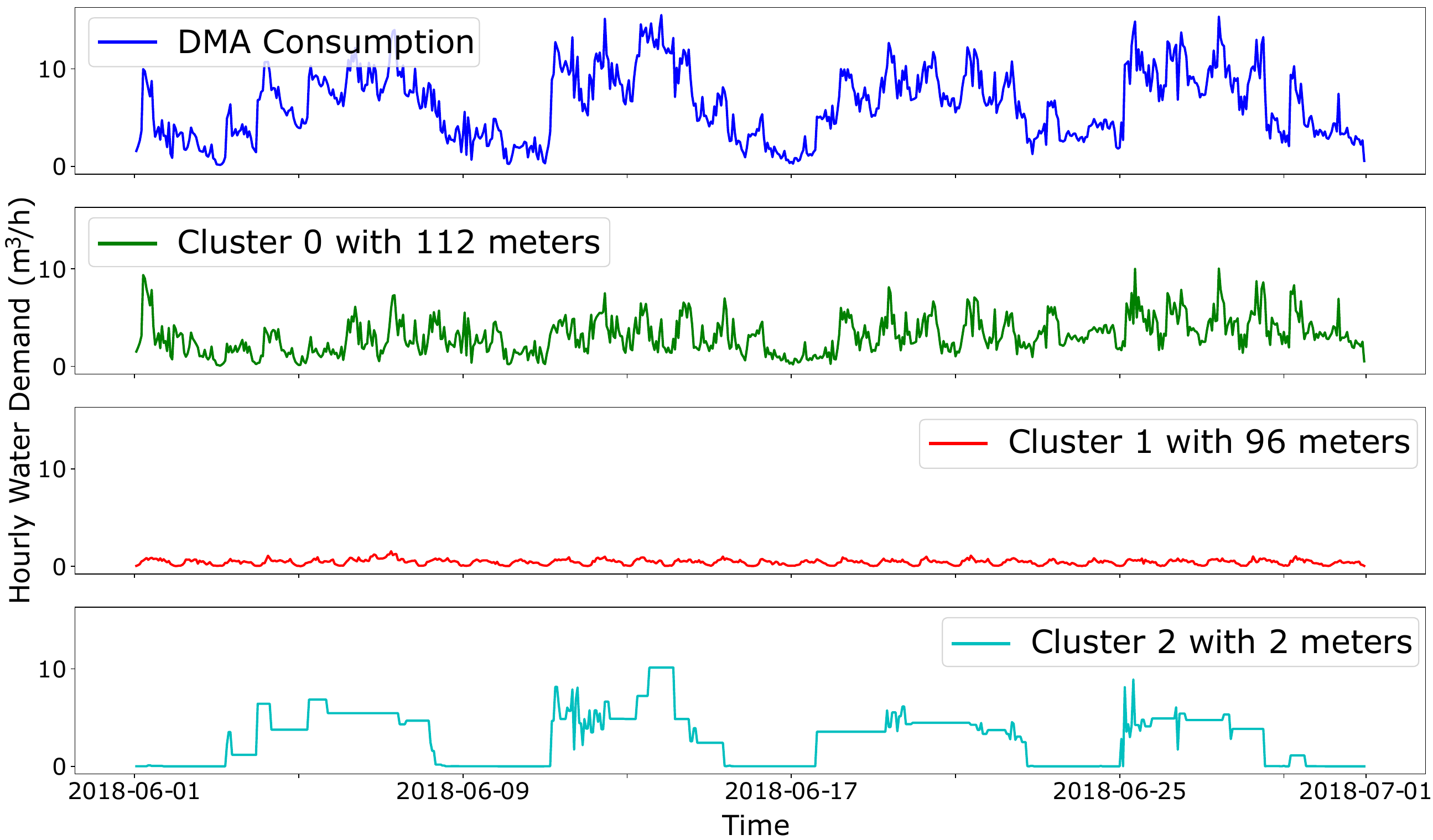}}
\caption{Observed consumption of \acrshort{dma} B along with the demand for the obtained clusters.}
\vspace{-14pt}
\label{Fig:cluster_b}
\end{figure}

The evaluation of the water end-use behaviors involves analyzing cluster formations across the \acrshort{dma}s, focusing on their distinct periodicity, trend, and consumption magnitude.
The results for \acrshort{dma} A and \acrshort{dma} B are depicted in Figure \ref{Fig:cluster_a} and \ref{Fig:cluster_b}, respectively, illustrating two contrasting scenarios: \acrshort{dma} A represents a heterogeneous consumer profile within a rural area, while \acrshort{dma} B exemplifies a heterogeneous profile comprising urban and agricultural area.
In all experiments, clustering through contrastive learning effectively separate varying end-user behaviors, thereby identifying the dominant demand patterns to enhance subsequent forecasting task.

In \acrshort{dma} A, a distinct long-term periodicity of approximately $45$ days is identified primarily in two end-consumers from cluster $\mathbf{0}$, whereas the remaining $157$ consumers demonstrate only short-term periodicity.
Further analysis, supported by the geolocation of smart meters and insights from the data provider's expert, reveals that this long-term periodicity within cluster $\mathbf{0}$ that has a significant difference to all other end-users is associated with a poultry farm, which typically has an average growing cycle of $6$ to $8$ weeks.

Conversely, in \acrshort{dma} B, the three distinct clusters underscore varied consumer behaviors within the \acrshort{dma}. Cluster $\mathbf{2}$, comprising two end-consumer meters linked to a corporate office, displays a consumption pattern aligned with typical workweek hours, exhibiting near-zero consumption from Friday evening to Monday morning. Clusters $\mathbf{0}$ and $\mathbf{1}$ demonstrate daily periodicity, albeit with differing consumption magnitudes, and are spatially dispersed across the \acrshort{dma}.
Cluster $\mathbf{1}$ predominantly includes small-scale commercial entities, whereas the remaining consumers in both clusters are primarily single-family residences with backyards. In contrast, \acrshort{dma} C lacks such dynamic consumer diversity, containing only two clusters with $14$ and $53$ meters respectively.

 \aboverulesep=0ex
 \belowrulesep=0ex

\begin{table}[h!]
    \centering
    \caption{Quantitative forecasting performance evaluation of the proposed approach, with baselines. The MAPE metric is \%, while MAE quantifies error in $l/h$.}
    \label{Tab:quant}
    \renewcommand{\arraystretch}{1.3} 
    \begin{tabular}{c|c|c|c|c|c}
        \toprule
        \textbf{Model} & \textbf{Metric} & \textbf{DMA A} & \textbf{DMA B} & \textbf{DMA C} & \textbf{Control} \\
        \midrule
        \multirow{2}{*}{ARIMA}  & MAPE  & $42.36$  & $92.05$  & $43.57$  & $59.67$  \\ \cmidrule(lr){2-6}
                                & MAE   & $50.38$  & $107.9$  & $67.53$  & $86.74$  \\
        \hline 
        \multirow{2}{*}{LSTM}   & MAPE  & $43.80$  & $82.41$  & $31.49$  & $12.98$  \\ \cmidrule(lr){2-6}
                                & MAE   & $9.25$   & $17.00$  & $7.48$   & $1.52$   \\
        \hline 

        \multirow{2}{*}{Wavelet CNN}  & MAPE  & $26.76$  & $41.85$  & $17.43$  & $8.48$   \\ \cmidrule(lr){2-6}
                                & MAE   & $6.34$   & $\mathbf{13.17}$  & $4.66$   & $0.97$   \\
        \hline 
        \multirow{2}{*}{Our Approach} & MAPE  & $\mathbf{21.86}$  & $\mathbf{40.27}$  & $\mathbf{16.95}$  & $\mathbf{8.40}$   \\ \cmidrule(lr){2-6}
                                  & MAE   & $\mathbf{5.40}$   & $13.52$  & $\mathbf{4.60}$   & $\mathbf{0.94}$   \\
        \bottomrule
    \end{tabular}%
\end{table}

The quantitative results for forecasting across all four \acrshort{dma}s are detailed in Table \ref{Tab:quant}.
Across experiments, our methods outperform the baseline LSTM method greatly, while introducing further improvement of different degrees compared to the Wavelet CNN baseline although the number of parameters are reduced by $97\%$.
Notably, \acrshort{dma} A, which exhibits two distinct periodicities, in terms of behavior and time frame, across its clusters, achieved the highest performance gain of $4.9\%$.
Conversely, \acrshort{dma} B and C, despite having clusters with varying behaviors, maintain weekly periodicities within each cluster, resulting in less pronounced differences compared to \acrshort{dma} A.
The control \acrshort{dma} illustrates the impact of statistical smoothing due to a large number of users, where the relative gain in forecasting performance from even 
the baseline LSTM is minimal, starkly contrasting with the significant improvements observed in \acrshort{dma}s A, B, and C. The statistical baseline of ARIMA exhibits high errors in absolute and relative terms, as it fails to predict both small and large deviations. 

\begin{figure}[bhtp]
\centerline{\includegraphics[width=3in,keepaspectratio=TRUE]{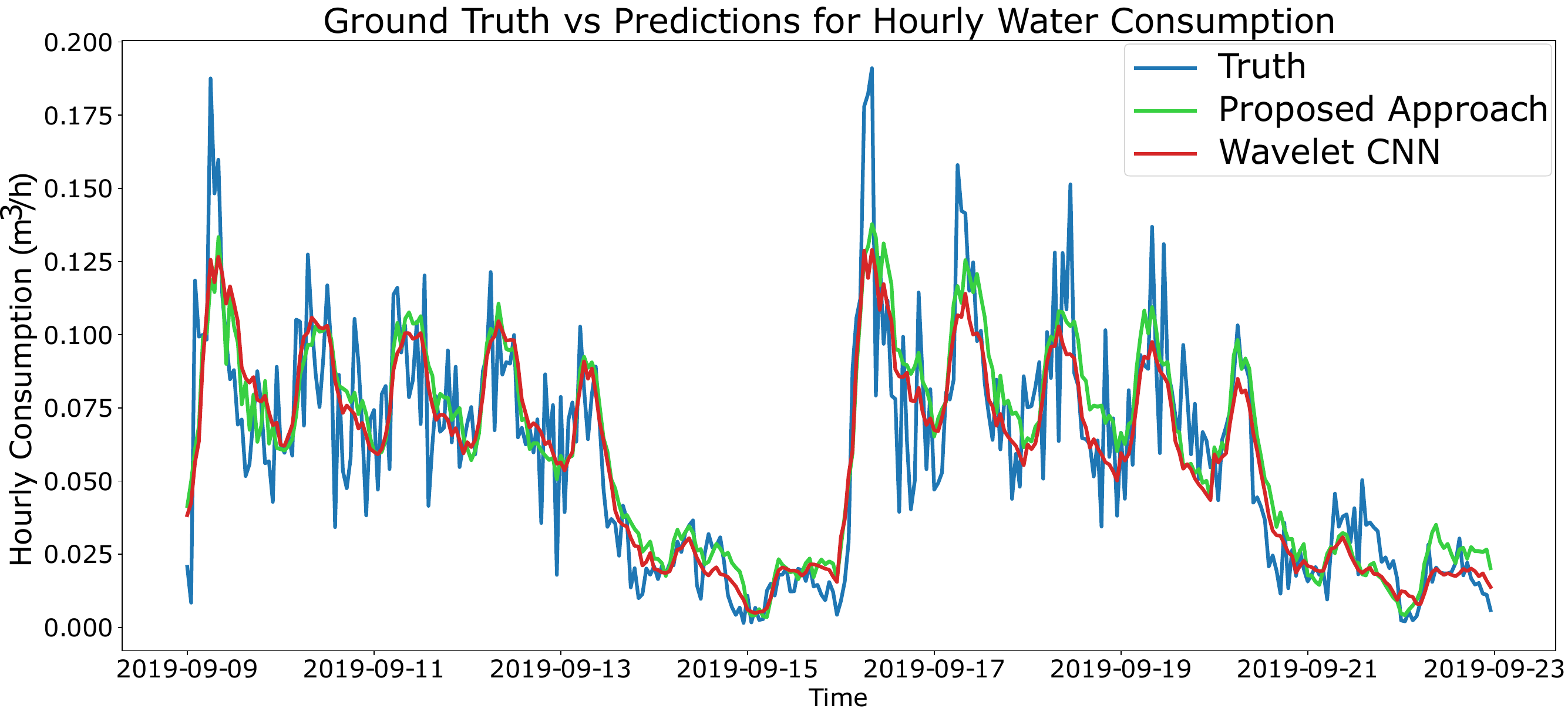}}
\caption{Qualitative comparison between observed and predicted water demand for \acrshort{dma} B over two weeks.}
\vspace{-10pt}
\label{Fig:qualitative}
\end{figure}

Further analysis of \acrshort{dma} B through a qualitative two-week plot (Monday to Sunday) shown in Figure \ref{Fig:qualitative} elucidates the reasons behind the low forecasting performance noted in Table \ref{Tab:quant}. All models primarily capture the trend and seasonality, struggling to account for the high variance, particularly evident during weekends and the sharp increase in demand on Mondays. This variance is attributed to the corporate end-consumer in cluster \textbf{2} of \acrshort{dma} B, as depicted in Figure \ref{Fig:cluster_b}. This effect significantly influences the models' forecasting capabilities on weekdays. Nonetheless, the proposed approach successfully identifies the end-user responsible for these stochastic effects during working days, providing critical insights for proactive water demand management.

%% file: Conclusion.tex
\section{Conclusion}

In this study, we propose and evaluate a novel framework for forecasting water demand in \acrshort{dma}s with heterogeneous end-consumers. Using contrastive learning, we identify user groups with varying behavioral patterns and enhance demand forecasts with features derived from these groups. We demonstrate this approach with two real-life examples of complex \acrshort{dma}s. We address the gap in modeling multi-faceted consumer demand beyond urban centers and elucidate the effect of statistical smoothing on large urban areas through a control group. Our method improves forecasting accuracy for \acrshort{dma} A by $4.9\%$, capturing both long- and short-term periodic patterns influenced by socio-economic factors. Additionally, clustering as a precursor for \acrshort{dma} B reveals end-consumers whose behavior obscures overall demand patterns, offering valuable insights for managing supply. As an extension, hybrid models combining deep learning and statistical methods could further improve forecasting for corporate consumers in \acrshort{dma} B, particularly for their piece-wise linear behavior during weekends. The effect of the proposed approach could be quantified on other architectures for time series forecasting.